\documentclass[10pt,twocolumn,letterpaper]{article}

\usepackage{iccv}
\usepackage{times}
\usepackage{epsfig}
\usepackage{graphicx}
\usepackage{amsmath}
\usepackage{amssymb}
\usepackage{subfigure}
\usepackage{multirow}
\usepackage{xcolor}
\usepackage{soul}

\usepackage[pagebackref=true,breaklinks=true,letterpaper=true,colorlinks,bookmarks=false]{hyperref}

\newcommand{\bd}[1]{\textbf{#1}}

\newcommand{\tablestyle}[2]{\setlength{\tabcolsep}{#1}\renewcommand{\arraystretch}{#2}\centering\footnotesize}
\makeatletter\renewcommand\paragraph{\@startsection{paragraph}{4}{\z@}
  {.5em \@plus1ex \@minus.2ex}{-.5em}{\normalfont\normalsize\bfseries}}\makeatother

\iccvfinalcopy 

\def\iccvPaperID{} 

\ificcvfinal\fi
\begin{document}

\title{Learning Joint 2D-3D Representations for Depth Completion}
\author{Yun Chen$^{1}$\quad Bin Yang$^{1,2}$\quad Ming Liang$^{1}$\quad Raquel Urtasun$^{1,2}$\\
{$^1$Uber Advanced Technologies Group\quad $^2$University of Toronto}
\\
{\tt\small \{yun.chen, byang10, ming.liang, urtasun\}@uber.com}
}

\maketitle
\thispagestyle{empty}

\begin{abstract}

In this paper, we tackle the problem of depth completion from RGBD data. 
Towards this goal, we design a simple yet effective neural network block that learns to extract joint 2D and 3D features.
Specifically, the block consists of two domain-specific sub-networks that apply 2D convolution on image pixels and continuous convolution on 3D points, with their output features fused in image space.
We build the depth completion network simply by stacking the proposed block, which has the advantage of learning hierarchical representations that are fully fused between 2D and 3D spaces at multiple levels.
We demonstrate the effectiveness of our approach on the challenging KITTI depth completion benchmark and show that our approach outperforms the state-of-the-art. 

\end{abstract}

\section{Introduction}

\begin{figure}[t]
 \centerline{\includegraphics[width=\linewidth]{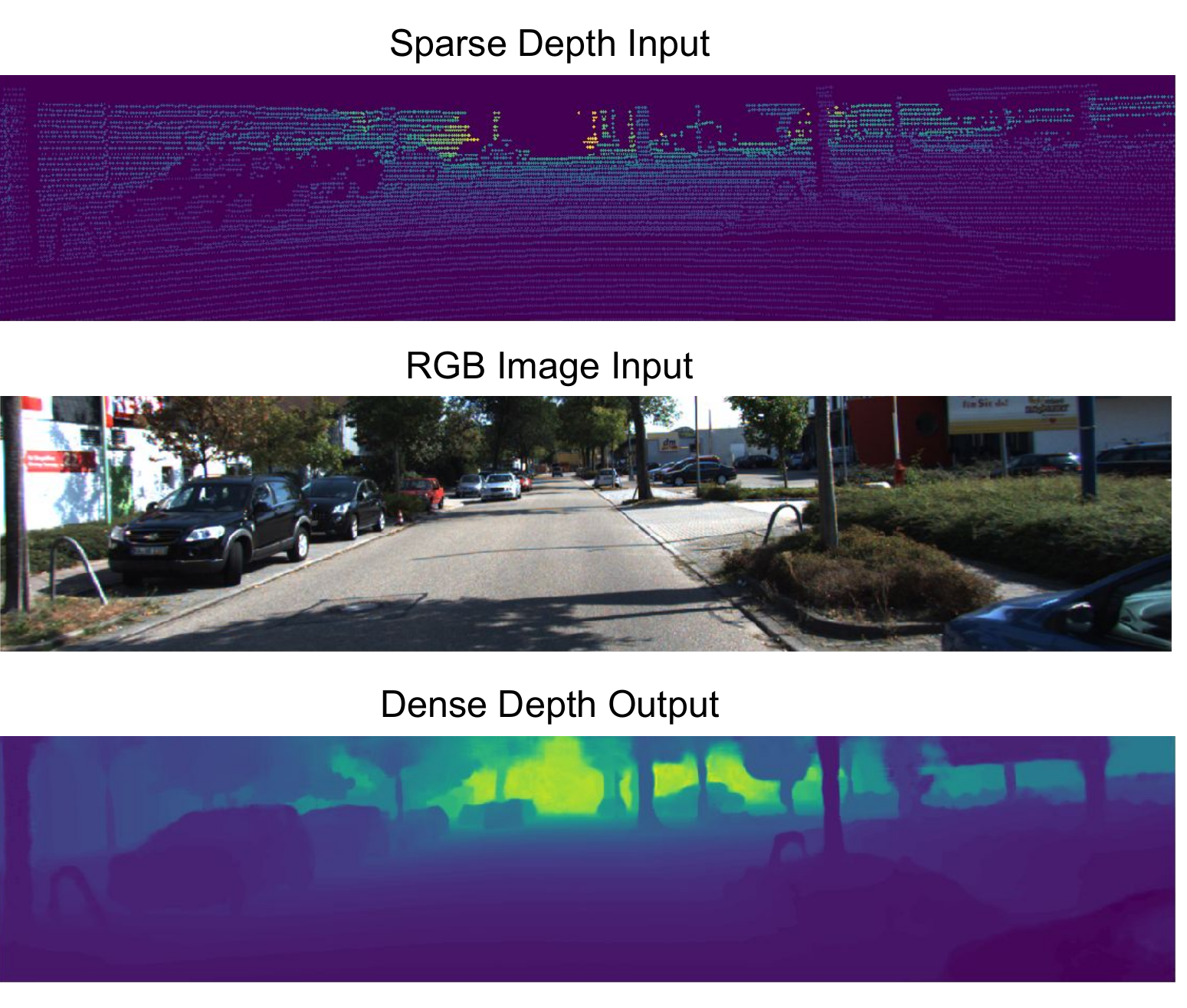}}
 \caption{ \textbf{Illustration of the depth completion task.} The model takes a sparse depth map (projection of the LiDAR point cloud) and a dense RGB image as input, and produces a dense depth map.}
 \label{fig:intro}
\end{figure}

In the past few years, the use of sensors that contain both image information as well as depth has increased significantly. They are typically used in applications such as self-driving vehicles, robotic manipulation as well as gaming. 
While passive sensors like cameras typically generate dense data, active sensors like LiDAR (Light Detection and Ranging) produce sparse depth observation of the environment. As a result, this semi-dense representation of the world can be inaccurate at regions close to object boundaries. One solution is to use high-end depth sensors with higher data density, but they are usually very expensive. A more affordable alternative is depth completion (shown in Figure \ref{fig:intro}), which takes the sparse depth observation and dense image as input, and estimates the dense depth map. In practice, depth completion is often employed as a precursor to downstream perception tasks such as detection, semantic segmentation or instance segmentation.

Despite many attempts to solve the problem, depth completion remains unsolved. 
Challenges such as the inherent ambiguity in extracting depth from images, as well as the noise and uncertainty in the unstructured sparse depth observation, make depth completion a non-trivial task. 

Many approaches \cite{sparsecnn,nconv,sparse2dense,dl,uncertainty} reason in the 2D space only by projecting the 3D point cloud to 2D image space. Convolutional neural networks (CNNs) are typically used to learn multi-modality representations in 2D space. However, as the metric space is distorted after the camera projection, such approaches have difficulty capturing precise 3D geometric clues. As a result, auxiliary task like surface normal estimation is added to better supervise the feature learning \cite{dl}. 
Other methods \cite{song2017semantic} reason in 3D space only by extracting 3D features (\eg Truncated Signed Distance Function \cite{tsdf}) from the sparse depth image of the scene and applies 3D CNN to learn 3D representations and complete the scene densely in 3D. The drawback is the lack of exploitation of the dense image data, which can provide discriminative appearance clues.

In contrast, in this paper, we take advantage of representations in both 2D and 3D spaces and design a simple yet effective architecture that fuses the information between these representations at multiple levels.
Specifically, we design a 2D-3D fuse block that takes feature map in 2D image space as input, branches into two sub-networks that learn representations in 2D and 3D spaces via multi-scale 2D convolutions and continuous convolutions \cite{contconv} respectively, and then fuses back into the 2D image space. Thanks to the modular design, we can create networks of various model sizes by simply stacking the 2D-3D fuse blocks sequentially.
Compared with other multi-sensor fusion based representations \cite{xu2018pointfusion, contfuse} that typically fuse the features from each sensor only once in the whole network, our proposed modular based model has the advantage of dense feature fusion at multiple levels through the network. As a result, while the domain-specific sub-networks inside the block extract specialized 2D and 3D representations separately, stacking such blocks together leads to hierarchical joint representation learning that fully exploits the complementary information between the two sensor modalities.

We validate our approach on the challenging KITTI depth completion benchmark \cite{sparsecnn}, and show that our approach outperforms all previous state-of-the-art methods in terms of Root Mean Square Error (RMSE) on depth. Note that our model is trained from scratch using KITTI training data only, and still surpasses other methods that exploit external data or multi-task learning. This further showcases the superiority of the proposed model in learning joint 2D-3D representations. We also conduct detailed ablation study to investigate the effect of each component of the model, and show that our model achieves better trade-off in accuracy versus model size compared with the state-of-the-art.

\section{Related Work}

In this section, we review previous literatures on the topics of depth estimation from RGB data, depth completion from RGBD data, and representation learning for RGBD data.

\subsection{Depth Estimation from RGB data}

Early approaches \cite{liu2014discrete,karsch2014depth,konrad20122d,saxena2006learning} estimated depth from single RGB images by applying probabilistic graphical models to hand-crafted features. With the recent advance in image recognition by deep convolutional neural networks (CNNs), CNN based methods are applied to depth estimation as well. Eigen \etal \cite{eigen2014depth} designed a multi-scale deep network for depth estimation from a single image. Laina \etal \cite{laina2016deeper} tackled the problem at a single scale by using a deep fully convolutional neural network. Liu \etal \cite{liu2016learning} combined deep representation with a continuous conditional random field (CRF) to get smoother estimations. Roy and Todorovic \cite{roy2016monocular} proposed to combine deep representations with random forests and achieved a good trade-off between prediction smoothness and efficiency. Recently unsupervised approaches \cite{garg2016unsupervised,godard2017unsupervised} exploited view synthesis as the supervisory signal, while some \cite{mahjourian2018unsupervised,wang2018learning,zhou2017unsupervised} further extended the idea to videos. However, due to the inherent ambiguity in depth from images, these approaches have difficulty producing high-quality dense depth.

\subsection{Depth Completion from RGBD data}

Different from depth estimation, the task of depth completion tries to exploit a sparse depth map (\eg point cloud scan from a LiDAR sensor) and possibly image data as well to predict high-resolution dense depth. Early work \cite{hawe2011dense,liu2015depth} resorted to wavelet analysis to generate dense depth/disparity from sparse samples. Recently, deep learning methods achieve superior performance in depth completion. Uhrig \etal \cite{sparsecnn} proposed sparse invariant CNNs to extract better representation from sparse input only. Ma \etal \cite{mal2018sparse} proposed to concatenate sparse depth together with RGB image and fed into an encoder-decoder based CNN for depth completion. A similar approach was also applied to the self-supervised setting \cite{sparse2dense}. Instead of using CNN, Cheng \etal \cite{cspn} used a recurrent convolution to estimate the affinity matrix for depth completion. Apart from the network architecture side, other methods exploited semantic contexts from multi-task learning. Schneider \etal \cite{schneider2016semantically} extracted object boundary cues for cleaner depth estimation. Semantic segmentation task was also exploited to jointly learn better semantic features of the scene \cite{spade, uncertainty}. Qiu \etal \cite{dl} added the auxiliary task of surface normal estimation to depth completion. Yang \etal \cite{ddp} learned a depth prior on images by training on large-scale simulation data. Compared with these approaches that focused on better network architecture and exploiting more context or prior from other dataset and labels. Our method improves performance simply by learning better representations. This is achieved by a new neural network block that's specially designed for RGBD data. We show in experiments that we are able to learn strong joint 2D-3D representations from the RGBD data with the proposed method and achieve state-of-the-art performance in depth completion. 

\subsection{Representation for RGBD data}

\begin{figure*}[tb]
 \centerline{\includegraphics[width=\linewidth]{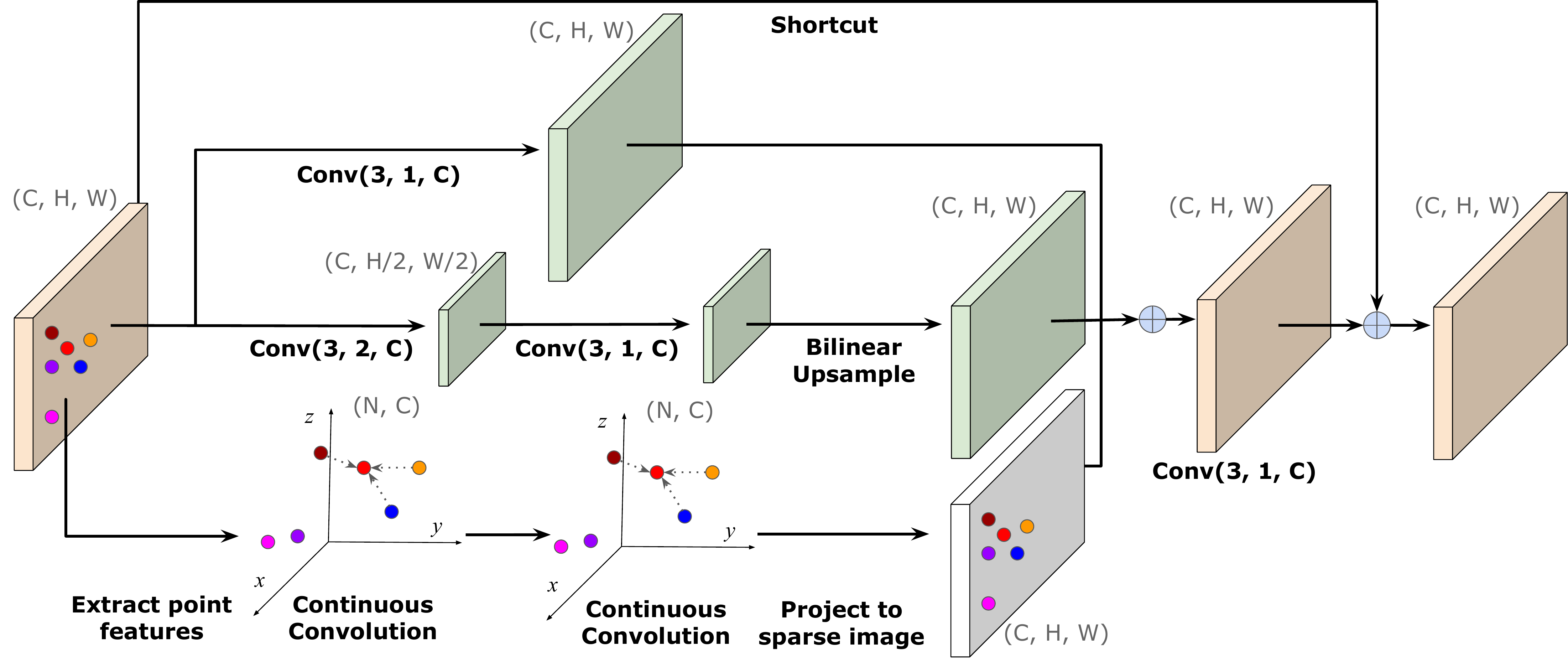}}
 \caption{\textbf{Architecture of the 2D-3D fuse block.} The 2D-3D fuse block consists of two branches, a multi-scale 2D convolution branch and a 3D continuous convolution branch. \texttt{Conv}$(k, s, c)$ denotes 2D convolution with kernel size $k$, stride $s$ and output channels $c$. The gray numbers in brackets denote the shape of features. The multi-scale 2D branch has two scales. One has the same scale as the input and is composed of one convolution. The other is downsampled by a stride 2 convolution, followed by a convolution and then bilinearly upsampled by 2. In the 3D branch, we first extract point features as the image features at the projection locations of the points, then apply two continuous convolutions, and finally project the points to image space to form a sparse image feature map. Continuous convolution uses K-Nearest-Neighbors algorithm to find the neighbors of each point. In the figure, we use K=3 as an example and only show the convolution operation on the red point. Note that the neighboring points in 2D space are not necessarily close to each other in 3D space. All convolutions are followed by batch normalization and ReLU.}
 \label{fig:in_block}
\end{figure*}

Song \etal \cite{song2014sliding} extracted multiple hand-crafted features (TSDF \cite{tsdf}, point density, 3D normal, 3D shape) from depth image for 3D object detection. In \cite{song2016deep} RGBD based joint representation was learned by applying 3D CNN to a 3D volume of depth image and 2D CNN to the RGB image and concatenating them together. Chen \etal \cite{mv3d} extracted 3D features by applying 2D CNN on multi-view projection of the 3D point cloud and combining with image features at ROI level. Xu \etal \cite{xu2018pointfusion} used the similar approach but adopted a PointNet \cite{qi2017pointnet} to extract 3D features on raw points directly. In \cite{wang2019densefusion} the same representation was further extended to pixel-level by fusing pixel feature with point feature. Liang \etal \cite{contfuse} first discretized the sparse LiDAR points into a dense bird's eye view voxel representation, and applied 2D CNN to extract BEV representations. The 2D image features are fused back to BEV space densely via continuous convolution \cite{contconv} to interpolate the sparse correspondence. Compared with these methods, our approach uses domain specific network for 2D and 3D representation learning, and both features are fused back to 2D image space at multiple levels across the whole network instead of only fusing once. As a result, we are able to learn more densely fused representation from the RGBD data.

\section{Learning Joint 2D-3D Representations}

We tackle the problem of depth completion from RGBD sensors. Existing approaches typically rely on either 2D or 3D representations to solve this task. 
In contrast, in this paper, we take advantage of both types of representations and design a simple yet effective architecture that fuses the information between these representations at multiple levels. 
In particular, we propose a new building block for neural networks that operates on RGBD data. It is composed of two branches that live in different metric spaces. In one branch we use traditional 2D convolutions to extract appearance features from dense pixels in 2D metric space. In the other branch, we use continuous convolutions \cite{contconv} to capture geometric dependencies from sparse points in 3D metric space. 
Our approach can be seen as spreading features to both 2D and 3D metric spaces, learning appearance and geometric features in each metric space separately, and then fusing them together.

We build our depth completion networks simply by stacking the 2D-3D fuse blocks. This modular design has two benefits. First, the network is able to learn joint 2D and 3D representations which are fully fused at multiple levels (all blocks). Second, the network architecture is simple and convenient to modify for the desired trade-off of performance and efficiency.

The remainder of the section is organized as follows: we first introduce our 2D-3D fuse block. We then give an example of deploying the proposed block to build a neural network for depth completion. Finally, we provide training and inference details of our depth completion network.

\subsection{2D-3D Fuse Block}

We show a diagram of the proposed 2D-3D fuse block in Figure \ref{fig:in_block}. The block takes as input a 2D feature map of shape $C \times H \times W$ and a set of 3D points of shape $N \times 3$. We assume that we are also given the projection matrix with which we can project the points from the 3D metric space to the 2D feature map. The output of the block is a 2D feature map with the same resolution as the input, which makes it straightforward to build a network by stacking the blocks for pixel-wise prediction tasks like depth completion. Inside the block, its architecture can be divided as two sub-networks: a multi-scale 2D convolution network and a 3D continuous convolution network. The input features are distributed to and processed in each sub-network, and their outputs are combined with a simple fusion layer. We refer readers to Figure \ref{fig:in_block} for an illustration of our method.
\begin{figure}[t]
 \centerline{\includegraphics[width=\linewidth]{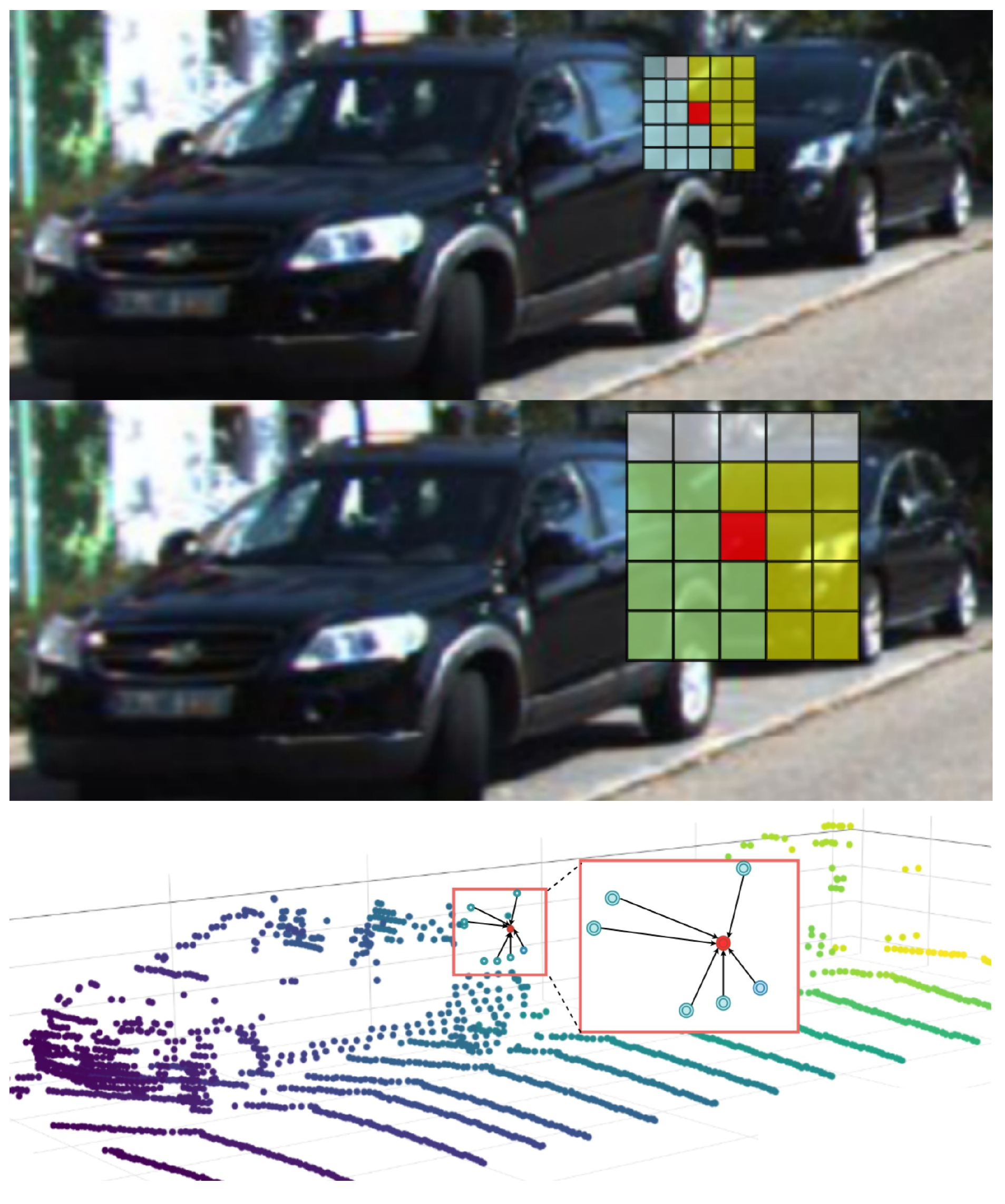}}
 \caption{ \textbf{Example receptive fields of \texttt{conv}($3$, $1$), \texttt{conv}($3$, $2$) and continuous convolution.} In 2D convolution, the neighbors are defined over image grids and are not necessarily close to each other in 3D space. The receptive field may cover both foreground and background objects. In the shown example convolution is performed at the red pixel. Green pixels are on the near car, and yellow pixels are on the distant car. In contrast, the neighbors in continuous convolution are based on the exact 3D geometric correlation.}
 \label{fig:receptive}
 \end{figure}

\paragraph{Multi-scale 2D convolution net:} We use a 2D convolution network to extract appearance features. We denote a 2D convolutional layer as \texttt{conv}($k$, $s$, $c$), where $k$ represents $k \times k$ filter size, $s$ denotes the convolution stride, and $c$ denotes the number of output channels. We adopt a two-branch network structure in order to extract multi-scale features. The first branch has the same resolution as the input and we simply apply \texttt{conv}(3, 1, $C$). The second branch consists of \texttt{conv}(3, 2, $C$), \texttt{conv}(3, 1, $C$) and \texttt{upsample}(2), where the first layer down-samples the feature map by $2$, and the last layer up-samples the feature map back to original resolution via bilinear interpolation. Batch normalization and ReLU non-linearity are used after each convolution. The outputs of both branches have the same shape $C \times H \times W$ as the input, and we combine them simply by element-wise summation.

\begin{figure*}[t]
    \centerline{\includegraphics[width=\linewidth]{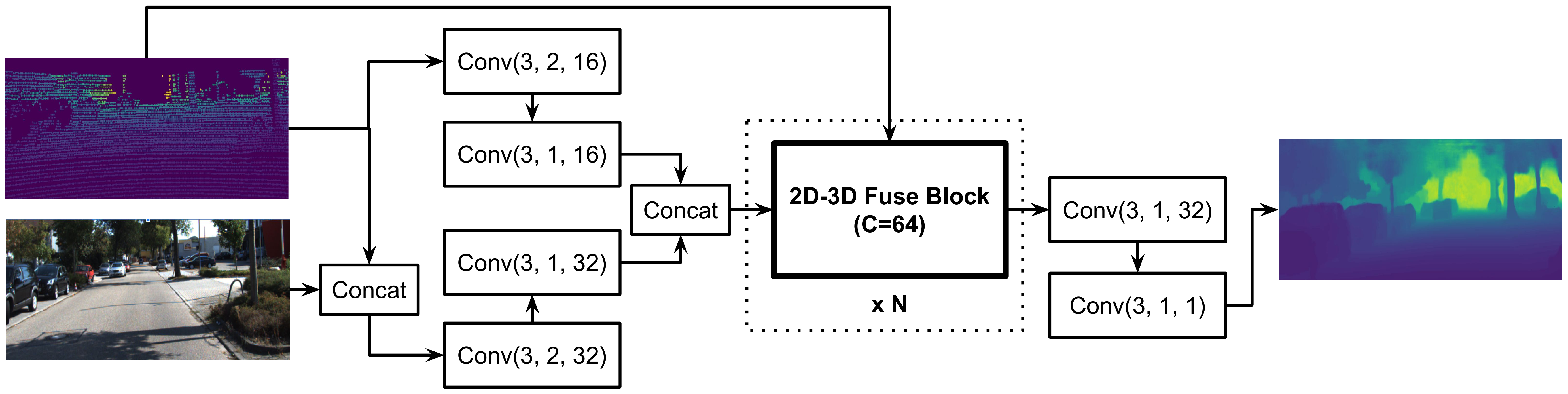}}
    \caption{ \textbf{Depth completion network based on 2D-3D fuse blocks.} The 2D-3D fused network takes image and sparse depth as input and predicts dense depth output. The main part of the network is the stacking of $N$ 2D-3D fuse blocks. We also apply some convolution layers at the input and the output stage.}
    \label{fig:in_net}
\end{figure*}

\paragraph{3D continuous convolution net:} We exploit continuous convolutions \cite{contconv} directly on the 3D points to learn geometric features in 3D metric space. The key concept of continuous convolution is the same as traditional 2D convolution, in that the output feature of each point is a weighted sum of transformed features of neighbors in a geometric space. But they use different ways to find neighbors and perform the weighted sum.
For 2D convolution the data is grid-structured so it is natural to use surrounding pixels as the neighbors of a center pixel. Moreover, each neighbor has its corresponding weight which is used to transform its feature before the summation. However, 3D points can be arbitrarily placed and their neighbors are not so natural as in grid data. In continuous convolution, we use K-Nearest-neighbors algorithms to find the $K$ neighbors of a point based on the Euclidean distance. We also parameterize the weighting function using a Multi-layer Perceptron (MLP). In practice, we use the following implementation of continuous convolution:
\begin{equation}
\mathbf{h}_i = W(\sum_{k} \text{MLP}(\mathbf{x}_i - \mathbf{x}_k) \odot \mathbf{f}_k)
\end{equation}
where $i$ is the index of points, $k$ is the index of neighbors, $\mathbf{x}$ denotes the 3-dimensional location of points, $\mathbf{f}_k$ and $\mathbf{h}_i$ denote the features, $W$ is a weight matrix, and $\odot$ denotes element-wise product. Note that the output of MLP has the shape as $\mathbf{f}_k$. This implementation can be regarded as a continuous version of separable convolution. The MLP and weighted sum perform depth-wise convolution, while the linear transformation resembles $1\times 1$ convolution. We make this separation to reduces the memory and computation overhead.

In our block, we first query the feature of each 3D point by projecting the point to the 2D feature map and extracting the feature at the projected pixel. After this step, we get 3D points of shape $N \times 3$ along with point features of shape $N \times C$. We then apply two continuous convolutions to the point feature. We use a two-layer MLP whose hidden feature dimension and output feature dimensions are $C/2$ and $C$ respectively. Each continuous convolution is followed by batch normalization and ReLU non-linearity. We then project the $N \times 3$ 3D points back to an empty 2D feature map and assign the $N\times C$ point features to corresponding projected pixels. In this way, we obtain a sparse 2D feature map as the output of the 3D sub-network. The output has the same shape as the outputs of the 2D sub-network.

\paragraph{Fusion:} Since the output feature maps of the 2D and 3D sub-networks have the same shape, we fuse them simply by element-wise summation. We then apply a \texttt{conv}(3, 1, $C$) layer to get the output of the 2D-3D fuse block. To facilitate training, we also add a shortcut connection from the input to the output when they have the same feature dimension.

Figure \ref{fig:receptive} illustrates the receptive field of 2D convolution and continuous convolution. While 2D convolution operates on neighboring pixels on grid-structured image feature maps, continuous convolution finds neighbors based on distance in 3D geometric space. By fusing the outputs of the two branches, the learned representation captures correlations in both spaces. At object boundaries, where depth estimation is usually hard for 2D convolution based methods, our approach has the potential to capture non-smooth representations for more accurate shape reconstruction by leveraging the geometric features in 3D space. We will show in experiments that our model predicts sharper and clear borders than other 2D representation methods.

\subsection{Stack 2D-3D Fuse Blocks into a Network}
Our 2D-3D fuse block can be used as a basic module to build the network. We simply stack a set of blocks plus a few convolution layers at the input and output stages to get our depth completion model. In Figure \ref{fig:in_net} we show the architecture of an example network with $N$ 2D-3D fuse blocks.

The inputs to the network include a depth image and an RGBD image. We first apply two convolution layers separately to each of the inputs. For the depth image, we use \texttt{conv}(3, 2, 16) and \texttt{conv}(3, 1, 16). For the RGBD image, we use \texttt{conv}(3, 2, 32) and \texttt{conv}(3, 1, 32). We then concatenate the two outputs and feed them to a stack of $N$ 2D-3D fuse blocks. The 3D points are obtained from the depth image and used by the blocks. We up-sample the output of the block set by 2 so that it has the same size as the input images. Finally, we apply another two convolution layers to obtain the output dense depth image. By stacking the blocks, the deep network is able to capture both large-scale context and local-scale clues, and the geometric and appearance features are fully fused in multiple levels.

\subsection{Learning and Inference}

We use a weighted sum of $\ell_2$ loss and smooth $\ell_1$ loss averaged over all image pixels that have depth labels as our default objective function.
\begin{equation}
\mathcal{L} = \ell_2 + \gamma \ell_1
\end{equation}
where $\gamma$ is the coefficient to control the balance between the two losses. The smooth $\ell_1$ loss of a pixel $i$ is defined as:
\begin{equation}
\ell_1(d_i, l_i) =
\begin{cases}
0.5(d_i-l_i)^2& \text{if } |d_i-l_i| < 1\\
|d_i-l_i| - 0.5& \text{otherwise},
\end{cases}
\end{equation}
where $d_i$ and $l_i$ are the predicted and ground truth depth, respectively.

Note that some other approaches use multi-task objective functions which leverage other tasks such as semantic segmentation to improve depth completion. Although we expect further performance gain with the multi-task objective function, we opt for the single task loss as the objective function is orthogonal to this work. During both training and inference, we pre-compute the indexes of nearest neighbors for all 3D points for continuous convolution, and apply the network to RGBD data and get the predicted results. No post-processing is required.

\section{Experimental Evaluation}

We conduct extensive experiments on KITTI depth completion benchmark \cite{sparsecnn} to validate the effectiveness of our approach.
Specifically, we compare with other depth completion methods on the test set by submitting to the KITTI evaluation server and show that our approach surpasses all previous state-of-the-art methods. We also conduct extensive ablation studies on the validation set to compare and analyze different model variants. Lastly, we provide some qualitative results of our approach.

\subsection{Experimental Setting}
\paragraph{Dataset:}
The KITTI depth completion benchmark \cite{sparsecnn} contains $86, 898$ frames for training, $1, 000$ frames for validation, and $1, 000$ frames for testing. Each frame has one sweep of LiDAR scan and an RGB image from the camera. The LiDAR and camera are calibrated already with the known transformation matrix. For each frame, a sparse depth image is generated by projecting the 3D LiDAR point cloud to the image. The ground-truth for depth completion is represented as a dense depth image, which is generated by accumulating multiple sweeps of LiDAR scans and projecting to the image. Note that depth outliers that are inconsistent with the stereo disparity label \cite{hirschmuller2008stereo} (caused by occlusion, dynamic objects or measurement artifacts) are removed from the ground-truth by ignoring the corresponding pixels during training and evaluation. We use both the RGB image and the sparse depth image as the input to our model.

\paragraph{Evaluation metrics:}
Four metrics are reported by the KITTI depth completion benchmark, which are Root Mean Square Error and Mean Absolute Error on depth (RMSE, MAE) and inverse depth (iRMSE, iMAE) respectively. We mainly focus on RMSE among all these metrics when comparing to other methods as it measures the error directly on depth and penalizes more on larger errors. The KITTI leaderboard also ranks methods based on RMSE. Additionally, we conduct an ablation study where we optimize the model with different objective functions and show that trade-off in different metrics can be controlled by different objective functions. Finding the best objective function for depth completion is out of the scope of this paper and we leave that for future work.

\paragraph{Implementation details:}
All images in KITTI validation and test sets are already cropped to the uniform size of $1216 \times 352$, while the training images are not. Therefore we randomly crop the training images (RGB, sparse depth and dense depth) to the size of $1216 \times 352$ during training. Thanks to the modular design of the proposed model, we can create different variants by changing the width (number of feature channels $C$) and depth (number of blocks $N$) of the network. For all model variants we initialize the network weights randomly, and train on 16 GPUs with a batch size of 32 frames. The training schedule goes as follows. We first train the model with $\ell_2$ loss for 100 epochs, with 0.0016 initial learning rate which is decayed by 0.1 at 65, 80, 85, 90 epochs respectively. We then fine-tune the model with the sum of $\ell_2$ and smooth $\ell_1$ loss for 50 epochs, with 0.00016 initial learning rate which is decayed by 0.1 at 30 epochs. In the 3D continuous convolution branch of the 2D-3D fuse block, we randomly sample 10, 000 points and apply a K-D tree to calculate the indices of 9 nearest neighbors and their relative distances for each point in advance.

\subsection{Comparison with State-of-the-art}

\begin{table}[t]
\setlength\tabcolsep{4pt}
\begin{center}
\begin{tabular}{l|c|c|c|c}
\hline
\multirow{2}{*}{Method} & \bd{RMSE} & MAE & iRMSE & iMAE \\
 & \bd{(mm)} & (mm) & (1/km) & (1/km) \\
\hline
SparseConvs \cite{sparsecnn} & 1601.33 & 481.27 & 4.94 & 1.78 \\
NN+CNN \cite{sparsecnn} & 1419.75 & 416.14 & 3.25 & 1.29 \\
MorphNet \cite{morphnet} & 1045.45 & 310.49 & 3.84 & 1.57 \\
CSPN \cite{cspn} & 1019.64 & 279.46 & 2.93 & 1.15 \\
Spade-RGBsD \cite{spade} & 917.64 & 234.81 & 2.17 & 0.95 \\
NConv-CNN-L1 \cite{nconv} & 859.22 & 207.77 & 2.52 & 0.92 \\
DDP$^\dagger$ \cite{ddp} & 832.94 & \bd{203.96} & \bd{2.10} & \bd{0.85} \\ 
NConv-CNN-L2 \cite{nconv} & 829.98 & 233.26 & 2.60 & 1.03 \\
Sparse2Dense \cite{sparse2dense} & 814.73 & 249.95 & 2.80 & 1.21 \\
DeepLiDAR$^\dagger$ \cite{dl} & 775.52 & 245.28 & 2.79 & 1.25 \\
FusionNet$^\dagger$ \cite{uncertainty} & 772.87 & 215.02 & 2.19 & 0.93 \\
\hline
Our FuseNet & \bd{752.88} & 221.19 & 2.34 & 1.14 \\
\hline
\end{tabular}
\end{center}
\caption{Comparison with state-of-the-art methods on the test set of KITTI depth completion benchmark, ranked by RMSE. $^\dagger$ indicates models trained with additional data and labels.}
\label{tab:test}
\end{table}

We evaluate our best single model on the KITTI test set, which has $N=12$ blocks stacked sequentially in the network, each with $C=64$ feature channels. We show the comparison results with other state-of-the-art methods on the KITTI depth completion benchmark in Table \ref{tab:test}. For a fair comparison, we mark methods that use external training data and labels in addition to KITTI training data. For example, DDP \cite{ddp} exploits the Virtual KITTI dataset \cite{vkitti} to learn the conditional prior of dense depth given an image. DeepLiDAR \cite{dl} pre-trains the model on the synthetic dataset generated from the CARLA simulator \cite{carla} to jointly learn the dense depth and surface normal tasks. FusionNet \cite{uncertainty} uses pre-trained semantic segmentation network on Cityscapes dataset \cite{cityscapes}. These methods rely on more data and various types of labels to learn good representations for depth completion. In contrast, our model, which is trained on KITTI training data only, outperforms all these methods considerably. This shows the superiority of the proposed model in learning joint 2D-3D representations from RGBD data over other methods. Specifically, our model significantly surpasses the second-best method with/without external data by 20/62 mm in RMSE respectively. We also achieve state-of-the-art results in other three metrics among methods that are trained on KITTI data only.

\subsection{Ablation Studies}

We conduct extensive ablation studies on the validation set of KITTI depth completion benchmark to justify the micro and macro design choices in the proposed model. We first compare different variants of the 2D-3D fuse block and then analyze the effect of different network configurations and objective functions. For faster experimentation, we conduct ablation studies on different network configurations with 100 training epochs only.

\paragraph{Receptive field of the continuous convolution branch:}
The proposed 2D-3D fuse block is composed of three branches, one 2D convolution branch, another 2D convolution branch with stride 2, and one 3D continuous convolution branch. Since we have varied the receptive fields of the 2D convolution by explicitly enumerating two different scales (stride 1 and stride 2), we wonder how to choose the receptive field of the 3D continuous convolution branch, which is controlled by the number of nearest neighbors. We show the ablation results in Table \ref{tab:knn}, where we can see that the model is quite robust to this hyper-parameter. In practice, we use $K=9$ nearest neighbors.

\paragraph{Architecture of the 2D-3D fuse block:}
We compare different architecture design of the 2D-3D fuse block in Table \ref{tab:ablation_block}. In particular, we want to know how much each convolution branch: the stride 1 and stride 2 2D convolutions and the continuous convolution, contributes to the final performance. As shown in Table \ref{tab:ablation_block}, multi-scale 2D convolution and continuous convolution are complementary. We rely on stride 1 convolution to extract the local features and continuous convolution to get 3D geometric features. Also, we need stride 2 convolution to extract better global features and propagate the sparse 3D geometric feature to a larger field. The results indicate that these three components are all necessary to the design of the 2D-3D fuse block for depth completion.


\begin{table}[t]
\begin{center}
\begin{tabular}{l|c|c|c|c|c}
\hline
K nearest neighbors & 3 & 6 & 9 & 12 & 15 \\
\hline
RMSE & 813 & \bd{810} & \bd{810} & 816 & 812 \\
\hline
\end{tabular}
\end{center}
\caption{Ablation study on number of nearest neighbors in the continuous convolution branch. Network config: $C=32, N=9$.}
\label{tab:knn}
\end{table}

\begin{table}[t]
\begin{center}
\begin{tabular}{c|c|c|c}
\hline
stride\_1 & stride\_2 & cont. & RMSE \\
conv & conv & conv & (mm)\\
\hline
& \checkmark & \checkmark & 840  \\ 
\checkmark &  & \checkmark &826  \\
\checkmark & \checkmark &  &817  \\
\checkmark& \checkmark & \checkmark &\textbf{803} \\
\hline
\end{tabular}
\end{center}
\caption{Ablation study on the architecture of the 2D-3D fuse block. Network config: $C=32, N=12$.}
\label{tab:ablation_block}
\end{table}

\begin{table}[t]
\begin{center}
\begin{tabular}{l|c|c|c|c}
\hline
Loss & RMSE & MAE & iRMSE & iMAE \\
\hline
$\ell_2$ & 790 & 232 & 2.51 & 1.16\\
smooth $\ell_1$ & 839 & \bd{197} & \bd{2.23} & \bd{0.91} \\
\hline
$\ell_2$, $\ell_2$ + smooth $\ell_1$ & \bd{785} & 217 & 2.36 & 1.08 \\
\hline
\end{tabular}
\end{center}
\caption{Ablation study on objective function. Network config: $C=64, N=12$.}
\label{tab:loss}
\end{table}

\begin{figure}[t]
 \includegraphics[width=1\linewidth]{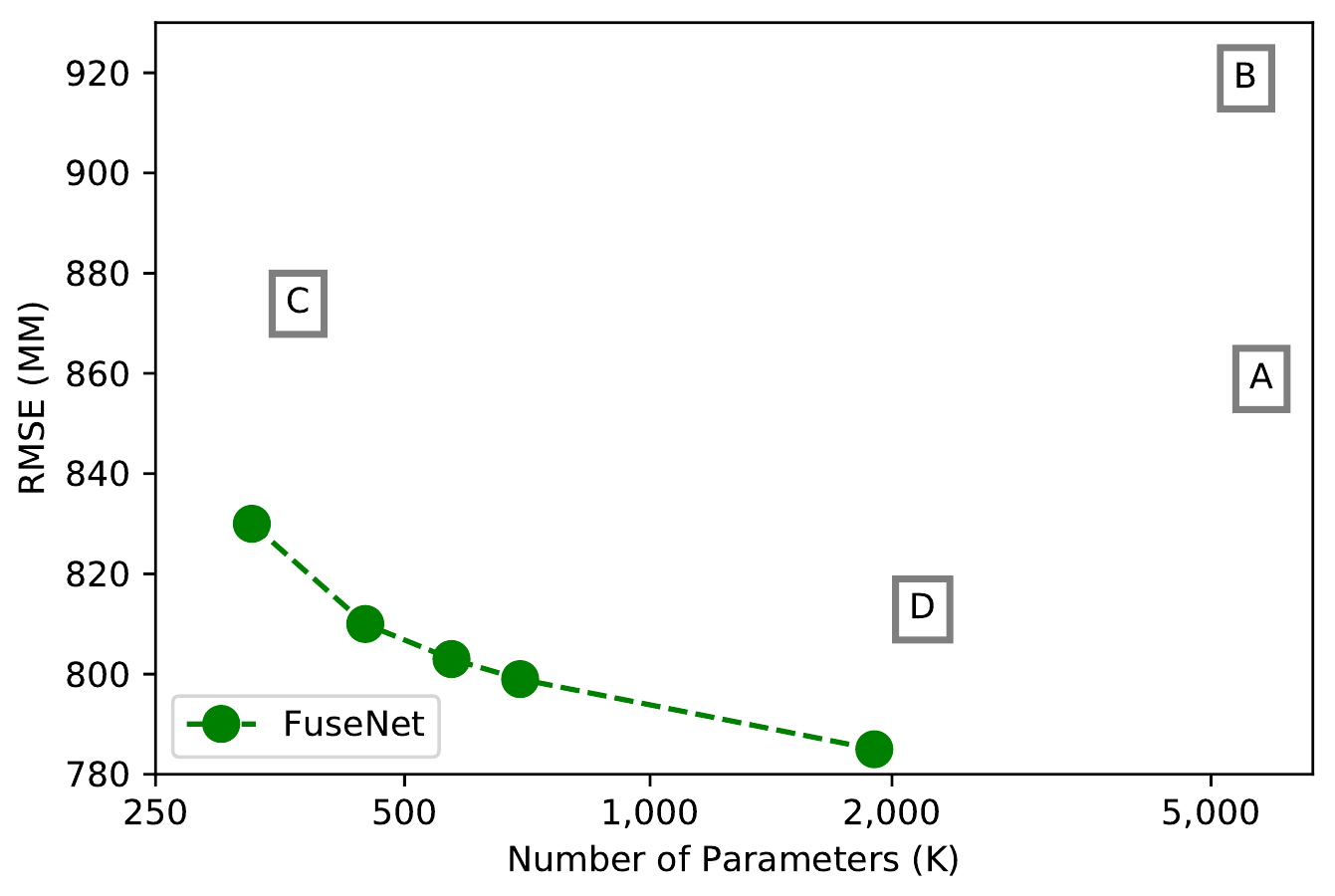}
 \hspace{-60mm}\resizebox{.55\columnwidth}{!}{\tablestyle{2pt}{1}
 \begin{tabular}[b]{lcc}
 Method & \#PARAM(K) & RMSE(MM) \\
 \hline
 \bd{[A]} Sparse2Dense \cite{sparse2dense} & 5540 & 857 \\
 \bd{[B]} Spade-RGBsD \cite{spade} & $\sim$5300 & 917 \\
 \bd{[C]} NConv-CNN-L2 \cite{nconv} & 355 & 872 \\
 \bd{[D]} FusionNet \cite{uncertainty} & 2091 & 811 \\
 \hline
 FuseNet-C32-N6 & \bd{322} & 830 \\
 FuseNet-C32-N9 & 445 & 810 \\
 FuseNet-C32-N12 & 568 & 803 \\
 FuseNet-C32-N15 & 692 & 799 \\
 FuseNet-C64-N12 & 1898 & \bd{785} \\
 \\
 \\
 \\
 \\
 \\
 \\
 \\
 \\
 \\
 \\
 \\
 \end{tabular}}\
\caption{Trade-off between accuracy and model size by varying feature channel number $C$ and block number $N$ of the network.}
\label{fig:tradeoff}
\end{figure}

\paragraph{Network configuration:}
We compare different network configuration by varying the width (number of feature channel $C$) and depth (number of blocks $N$) of the network. As a result, we are able to achieve different trade-offs between performance and model size. We plot the results in comparison with other methods in Figure \ref{fig:tradeoff}, where we show that our model achieves better performance with a smaller model size compared with other methods.

\paragraph{Objective function:}
We note that performance on different metrics can be controlled by employing different loss functions. Intuitively better RMSE metric could be achieved by $\ell_2$ loss, while better MAE metric could be achieved by $\ell_1$ loss. We validate this by comparing models trained with $\ell_2$ loss and smooth $\ell_1$ loss respectively for 100 epochs. The results are shown in Table \ref{tab:loss}. To get a better balance on all four metrics, our best single model is trained with $\ell_2$ loss for 100 epochs first and then trained with the sum of $\ell_2$ and smooth $\ell_1$ loss for another 50 epochs.

\begin{figure*}[t]
 \centering
 \begin{tabular}{*{3}{c@{\hspace{3.5px}}}}
 \multicolumn{3}{c}{\includegraphics[width=0.99\textwidth]{{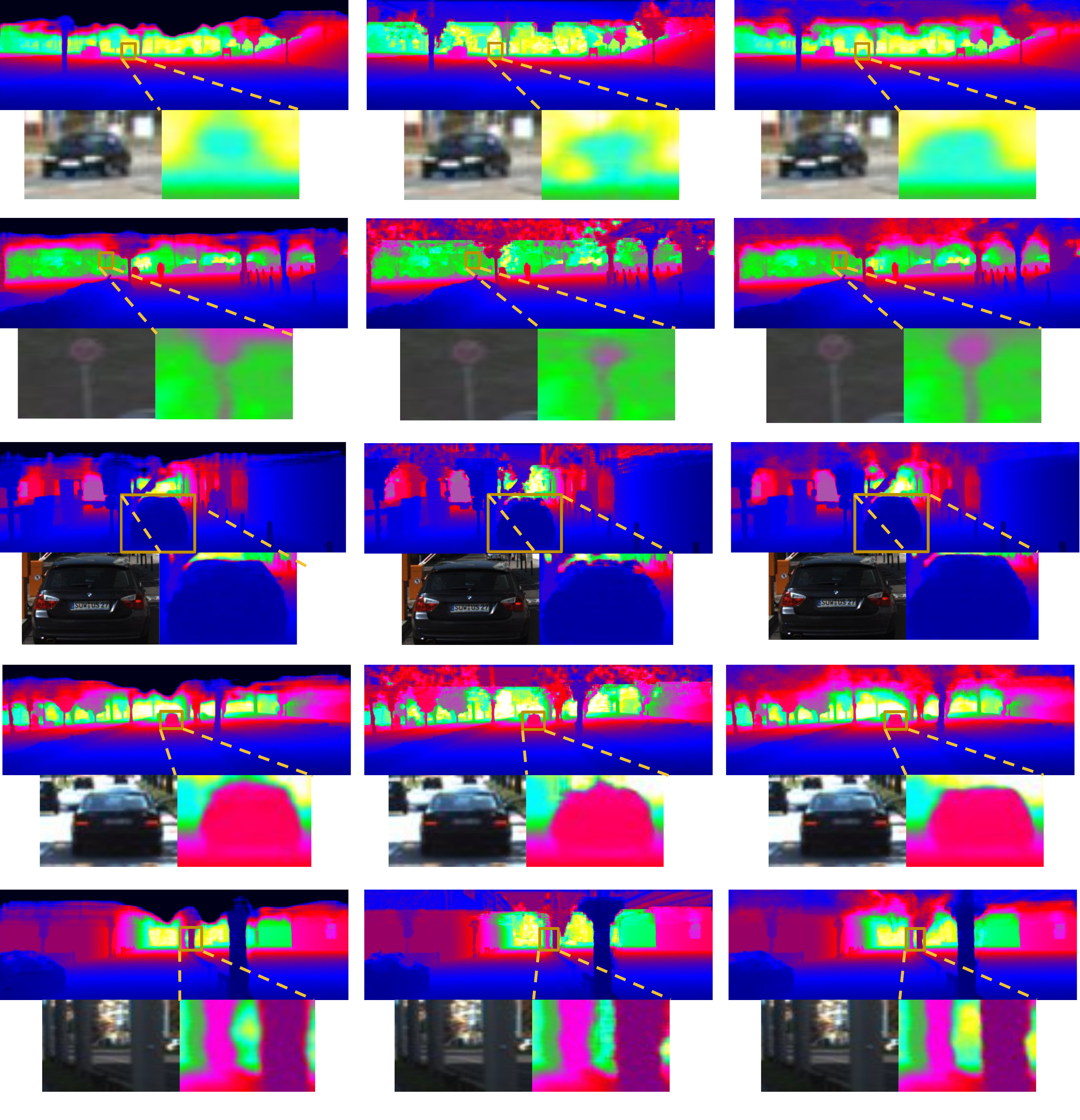}}}
 \\
 \hspace{20pt} \small{ 
 \hspace{20pt}Sparse2Dense \cite{sparse2dense}} \hspace{90pt} & \small{ NConv-CNN-L2 \cite{nconv}} \hspace{30pt} & \small{Ours} \hspace{30pt}\\
 \\
 \end{tabular}
 \caption{Qualitative results in comparison with two state-of-the-art methods (better viewed in color). Our model produces sharper boundaries of objects especially in the long range.}
 \label{fig:demo}
\end{figure*}

\subsection{Qualitative Results}
We show some qualitative results of the proposed method in comparison with two state-of-the-art methods NConv-CNN \cite{nconv} and Sparse2Dense \cite{sparse2dense} on the test set of KITTI depth completion benchmark. As shown in Figure \ref{fig:demo}, due to the use of continuous convolution that captures accurate 3D geometric features, our approach produces cleaner and sharper object boundaries in both near and distant regions. We get significantly better results for distant objects where 2D convolution can barely handle due to limited appearance clues. This suggests that in the task of depth completion, the description of the scale-invariant geometric feature in 3D is very important, and the proposed 2D-3D fuse block provides a simple yet effective solution to learn joint 2D and 3D representations.
\section{Conclusion}

In this paper, we have proposed a simple yet effective architecture that fuses information between 2D and 3D representations at multiple levels. We have demonstrated the effectiveness of our approach on the challenging KITTI depth completion benchmark and show that our approach outperforms the state-of-the-art. 
In the future, we plan to extend our approach to fuse other sensors and reason about video sequences.

{\small
\bibliographystyle{ieee_fullname}
\bibliography{egbib}
}

\end{document}